\documentclass[10pt,twocolumn,letterpaper]{article}

\usepackage{cvpr}              

\usepackage{graphicx}
\usepackage{amsmath}
\usepackage{amssymb}
\usepackage{booktabs}
\usepackage{tabularray}
\usepackage{multirow}

\usepackage[pagebackref,breaklinks,colorlinks]{hyperref}

\usepackage[capitalize]{cleveref}
\crefname{section}{Sec.}{Secs.}
\Crefname{section}{Section}{Sections}
\Crefname{table}{Table}{Tables}
\crefname{table}{Tab.}{Tabs.}

\def\cvprPaperID{*****} 
\def\confName{CVPR}
\def\confYear{2022}

\newcommand\blfootnote[1]{%
  \begingroup
  \renewcommand\thefootnote{}\footnote{#1}%
  \addtocounter{footnote}{-1}%
  \endgroup
}

\begin{document}

\title{EgoVideo: Exploring Egocentric Foundation Model and Downstream Adaptation}

\author{
    Baoqi Pei$^{1,2*}$,
    Guo Chen$^{3,1*}$, 
    Jilan Xu$^{4,1*}$,
    Yuping He$^{3*}$,
    Yicheng Liu$^{3*}$,
    Kanghua Pan$^{3*}$,\\
    Yifei Huang$^{1,5*}$, 
    Yali Wang$^{1,6}$,
    Tong Lu$^{3}$,
    Limin Wang$^{1,3}$,
    Yu Qiao$^{1}$ \\ \\
    $^1$Shanghai AI Laboratory,
    $^2$Zhejiang University,
    $^3$Nanjing University,\\
    $^4$Fudan University,
    $^5$The University of Tokyo, 
    $^6$SIAT, CAS\\
    {\tt\small peibaoqi@gmail.com}~~~~
    {\tt\small jilanxu18@fudan.edu.cn}~~~~
    {\tt\small chenguo1177@gmail.com}\\
    {\tt\small \{502023330020,522023330056,522023330071\}@smail.nju.edu.cn}~~~~
    {\tt\small hyf015@gmail.com}\\
    {\tt\small yl.wang@siat.ac.cn}~~~~
    {\tt\small \{lutong,lmwang\}@nju.edu.cn}~~~~
    {\tt\small qiaoyu@pjlab.org.cn}
}
\maketitle
\blfootnote{$*$ These authors contributed equally.}
\begin{abstract}
In this report, we present our solutions to the EgoVis Challenges in CVPR 2024, including five tracks in the Ego4D challenge and three tracks in the EPIC-Kitchens challenge. 
Building upon the video-language two-tower model and leveraging our meticulously organized egocentric video data, we introduce a novel foundation model called \textbf{EgoVideo}. This model is specifically designed to cater to the unique characteristics of egocentric videos and provides strong support for our competition submissions.
In the Ego4D challenges, we tackle various tasks including Natural Language Queries, Step Grounding, Moment Queries, Short-term Object Interaction Anticipation, and Long-term Action Anticipation. 
In addition, we also participate in the EPIC-Kitchens challenge, where we engage in the Action Recognition, Multiple Instance Retrieval, and Domain Adaptation for Action Recognition tracks. By adapting EgoVideo to these diverse tasks, we showcase its versatility and effectiveness in different egocentric video analysis scenarios, demonstrating the powerful representation ability of \textbf{EgoVideo} as an egocentric foundation model. Our codebase and pretrained models are publicly available at \url{https://github.com/OpenGVLab/EgoVideo}.
\end{abstract}

\section{Introduction}

In computer vision research, egocentric video understanding represents a pivotal task aimed at enabling machines to comprehend videos including human activities from a first-person perspective. Unlike traditional third-person viewpoint analysis, egocentric video understanding focuses on understanding human activities as they occur from the camera wearer's viewpoint, often captured through wearable cameras or head-mounted devices. This task holds significant implications across various domains, including healthcare~\cite{zhang2022seeing}, virtual/augmented reality~\cite{zhao2024instance}, and human-computer interaction~\cite{damen2014you}. 
Egocentric video action understanding facilitates applications ranging from assistive technologies for the visually impaired to immersive experiences in virtual environments~\cite{chalasani2018egocentric,huang2018predicting}. 
Additionally, it fosters advancements in personalized assistance systems~\cite{yao2019egocentric}, sports analytics~\cite{aghdam2015unsupervised}, and surveillance technologies~\cite{de2009egocentric}, thereby underscoring its multifaceted impact on both academic research and practical applications.

In recent years, action recognition methods have undergone significant advancements, propelled by the surge in deep learning techniques and the availability of large-scale annotated datasets~\cite{k400,ego4d,ek100}. 
With the advent of convolutional neural networks (CNNs)~\cite{c3d,k400} and recurrent neural networks (RNNs), action recognition has witnessed a paradigm shift towards end-to-end trainable models capable of automatically learning discriminative features from video clips. Furthermore, the integration of attention mechanisms, spatial-temporal modeling, and graph-based representations has further enhanced the performance of action recognition systems. 
Benefit from large-scale vision-language datasets~\cite{webvid,internvid,ego4d}, a variety of video foundation models~\cite{internvideo2,videococa} have been designed to learn general video representations, which have shown to benefit a series of downstream action recognition tasks~\cite{k400,sthv2,ek100}. However, as most of these video foundation models are trained on videos recorded in third-person view, the learned representations turn out sub-optimal for egocentric video understanding~\cite{egoinstructor,egoexolearn,egoexo4d,yang2022interact,huang2020improving}.

\begin{figure*}
    \centering
    \includegraphics[width=\linewidth]{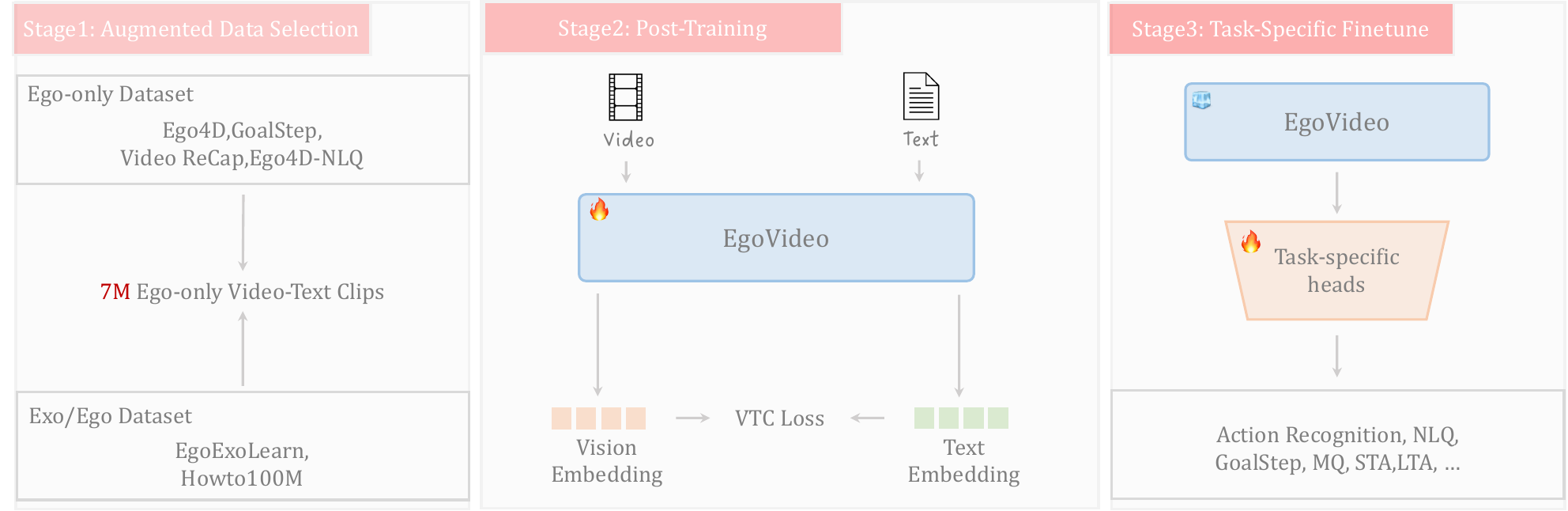}
    \caption{The workflow of the training process of EgoVideo. It includes 3 stages: in the first stage, we filter and select high-quality egocentric video and text pairs from multiple existing datasets. Then we perform post pertaining using the data in stage 1 by standard video-text contrastive learning. Finally, we adapt the pretrained EgoVideo model to different downstream tasks.}
    \label{fig:workflow}
\end{figure*}

To tackle the challenge, we propose a 3-stage training paradigm for egocentric video understanding, including multiple tasks like natural language grounding, domain adaptation, and multi-instance retrieval. Specifically, we first filter and select high-quality egocentric video and text pairs from multiple existing datasets~\cite{ego4d,howto100m,goalstep,egoexolearn}. These high-quality data serve as the foundational data for transferring models learned from general domains to the egocentric domain.
We adopt a video foundation model~\cite{internvideo2} that is pre-trained on large-scale video-language datasets~\cite{internvid}. 
With the help of rich vision features and a wide range of action-aware knowledge, this model is capable of extracting general video feature representations, acting as a good starting point for subsequent feature learning. 
In the second stage, to mitigate the domain gap between web-scale video datasets and egocentric videos, we perform post-training on the selected data, effectively transferring the general video feature representations to egocentric domain. 
We term the resulting model as \textbf{EgoVideo}, consisting of a strong egocentric video encoder \textbf{EgoVideo-V} and a text encoder \textbf{EgoVideo-T}.  
In the third stage, we conduct task-specific fine-tuning of EgoVideo-V and EgoVideo-T on three different egocentric video understanding tasks, \emph{e.g.}, natural language queries, domain adaptation action recognition, and multi-instance retrieval. 

Experimental results show that our 3-stage strategy has led to a remarkable improvement in overall model performance. The model excels at understanding fine-grained, action-specific information, demonstrating strong performance in action recognition and multi-instance retrieval. Moreover, benefiting from the multi-stage training, our model exhibits video understanding ability across a wide range of actions.

In the remainder of this report, we will detail our solutions along with experiments for each joined Ego4D track. Finally, we discuss the limitations of our work and conclude this technical report.

\section{Training Process of EgoVideo}

\subsection{Stage1: Augmented Data Selection}
To better transfer the video foundation model learned in the general video domain into the egocentric domain, we collect a broad range of paired egocentric video-text pairs from public video datasets, such as Ego4d~\cite{ego4d}, HowTo100M~\cite{howto100m}, EgoExoLearn~\cite{egoexolearn}, and Ego4d GoalStep~\cite{goalstep} by automatic filtering techniques. We do this to ensure a wider range of egocentric data and maintain the pertaining data quality. This results in around 7M video-text pairs. 
  
\subsection{Stage2: Egocentric Video Post-training}
In this work, we adopt InternVideo2~\cite{internvideo2}, a novel video foundation model that is pre-trained on millions of video-text pairs~\cite{internvid}. 
InternVideo2 is built through a progressive learning scheme, consisting of feature distillation, multi-modal alignment, and vision-language connection.
The pre-trained video foundation model thus acts as a strong starting point for the subsequent feature learning process. More details about the foundation model can be found in ~\cite{internvideo2,mvbench}.

We then perform the post-pretraining process and train the model for 5 epochs on the hybrid data in Stage 1 to improve the egocentric video understanding ability.
The model is optimized via a standard visual-text contrastive loss. 
During training, we also examine the model's egocentric video understanding ability on EPIC-Kitchen-100 zero-shot multi-instance retrieval benchmark~\cite{ek100}, and the results are shown in Table~\ref{tab:zeroshot}. 
We term this egocentric video foundation model as \textbf{EgoVideo}, consisting of a strong egocentric video encoder \textbf{EgoVideo-V} and a text encoder \textbf{EgoVideo-T}.

\subsection{Stage 3: Egocentric Downstream Adaptation}
After stage 2 training, we obtain a video foundation model EgoVideo tailored for the egocentric domain. We use this model to initialize the models in stage 3. In this stage, we conduct task-specific fine-tuning on the training sets. We put the detailed task-specific fine-tuning process of each task in the following section.

\section{Task-specific Finetuning}
\subsection{Task 1: Natural Language Queries @ Ego4D}

\noindent\textbf{Task Definition}
Given a video clip and a natural language query, the Ego4D \cite{ego4d} Natural Language Queries task aims to identify the temporal window corresponding to the query's answer.

\noindent\textbf{Approach}
Our solution builds upon GroundNLQ\cite{hou2023groundnlq} and employs our EgoVideo to extract video and text features. 
GroundNLQ proposes a multi-modal multiscale transformer encoder module to encode both video and text features and then efficiently fuse them. Following GroundNLQ, we first pretrain on NaQ~\cite{ramakrishnan2023naq} data and then fine-tune on NLQ data. The key driver for this task is Yuping He.

\noindent\textbf{Implementation Details.}
\emph{1) Feature Extraction:} We leverage ViT-1B of EgoVideo to extract video feature for each snippet, which contains $s=16$ consecutive frames with interval $\delta=16$. The text features are extracted by BERT-Large of EgoVideo.
\emph{2) Training Setup:} In the pretraining phase, we set the batch size to 8 and the total epochs to 10, with a warmup of 4 epochs, employing a maximum learning rate of 2e-4. 
In the fine-tuning phase, we set the batch size to 2 and the total epochs to 10, with a warmup of 4 epochs, with a maximum learning rate of 5e-5. 

\begin{table*}[h]
\centering
\small
\setlength\tabcolsep{1.6mm}
\begin{tabular}{lllcccccccc}
\toprule
\multirow{3}{*}{\#} &\multirow{3}{*}{Method} & \multirow{3}{*}{Feature}& \multicolumn{4}{c}{Validation} & \multicolumn{4}{c}{Test} \\
 \cmidrule(lr){4-7} \cmidrule(lr){8-11}
& & & R1@0.3 &R1@0.5 & R5@0.3 & R5@0.5 & R1@0.3 &R1@0.5 & R5@0.3 & R5@0.5 \\
\midrule 
A&GroundNLQ~\cite{hou2023groundnlq} & EgoVLP &  26.98 & 18.83 & 53.56 & 40.00 &  24.50 & 17.31 & 40.46 & 29.17\\
B&GroundNLQ\dag ~\cite{hou2023groundnlq} & EgoVLP & 27.20 & 18.91 & 54.42 & 39.98 & 25.67 & 18.18 & 42.05 & 29.80  \\ 
C&GroundVQA~\cite{di2023grounded}& EgoVLP & 29.70 & - & - & - & 26.67 & 17.63 & 39.94 & 27.70 \\ 
D&GroundNLQ & EgoVideo & 28.65 & 19.73 & 53.30 & 40.42 & 25.07 & 17.31 & 40.88 & 29.67 \\ 
\textbf{E}&\textbf{C+D} & \textbf{Ensemble} & - & - & - & - & \textbf{28.05} & \textbf{19.31} & \textbf{44.16} & \textbf{31.37} \\ 
\bottomrule
\end{tabular}
\caption{The Natural Language Queries performance.  }
\label{table:result}
\end{table*}

\noindent\textbf{Results.}
Table~\ref{table:result} presents the results of NLQ. \#A and \#D employ identical model and training strategy, while our single model’s features( \#D) significantly outperform the ensemble of EgoVLP and InternVideo(\#A). \#E combines predictions from GroundNLQ, GroundNLQ*, and GroundVQA~\cite{di2023grounded}. GroundNLQ* is a variant of GroundNLQ, distinguished by the integration of a cross-modal layer within the encoder. GroundVQA leverages a large language model to encode visual and language features. Ensemble methods further enhance performance.

\begin{table}[t]
\centering
\small
\setlength\tabcolsep{1.6mm}
\begin{tabular}{lcccc}
\toprule
 & \multicolumn{2}{c}{Validation} & \multicolumn{2}{c}{Test} \\
  \cmidrule(lr){2-3} \cmidrule(lr){4-5}
Method & R1@0.3 &R1@0.5 & R1@0.3 & R1@0.5 \\

\midrule 
VSLNet~\cite{vslnet} &  - & - & 19.04 & 12.04 \\
Ours & 28.02 & 23.66 & 32.99 & 25.92\\ 
\textbf{Ours (Ensemble)} & - & - & \textbf{34.06} & \textbf{26.97}\\ 
\bottomrule
\end{tabular}
\caption{The Step Grounding performance. Our method is the combination of GroundNLQ and EgoVideo.}
\label{table:goalstep_result}
\end{table}

\subsection{Task 2: GoalStep - Step Grounding @ Ego4D}

\noindent\textbf{Task Definition}
Step grounding aims to identify the temporal segment in an untrimmed egocentric video corresponding to a given natural language description of the step.

\noindent\textbf{Approach}
Similar to NLQ, we use GroundNLQ as the grounding model for Step Grounding and adopts EgoVideo to extract video and text features. The key driver for this task is Yuping He.

\noindent\textbf{Implementation Details.} We adopt the consistent configurations with NLQ for feature extraction.
During the fine-tuning phase, we use a batch size of 8, apply dropout with a probability of 0.2, and set the drop path rate to 0.2. Other hyperparameters remain the same as in NLQ.

\noindent\textbf{Results.}
Table~\ref{table:goalstep_result} displays our results on Step-Grounding. The official baseline uses VSLNet\cite{vslnet} as the grounding model and Omnivore\cite{girdhar2022omnivore} features. In contrast, our solution leverages stronger video and text features along with advanced grounding models, resulting in notable improvements. After ensembling results from GroundNLQ and GroundNLQ$\ast$, we achieve further gains.

\subsection{Task 3: Moment Queries @ Ego4D}

\noindent\textbf{Task Definition}
Given an egocentric video and a specific action category, Moment Queries task aims to retrieve all temporal segments corresponding to this action category. The action categories are pre-defined and specific to first-person activities.

\begin{table*}[t]
\centering
\small
\setlength\tabcolsep{4.6mm}
\begin{tabular}{llcccc}
\toprule
\multirow{3}{*}{\#} & \multirow{3}{*}{Feature} & \multicolumn{2}{c}{Validation} & \multicolumn{2}{c}{Test} \\
  \cmidrule(lr){3-4} \cmidrule(lr){5-6}
&  & Average mAP & R1@0.5 & Average mAP &R1@0.5 \\
\midrule 
A&InternVideo + EgoVLP & 27.85 & 46.98 & - & - \\
B&Slowfast + Omnivore + EgoVLP + InternVideo &  - & - & 29.34 & 48.50\\
C & \textbf{EgoVideo-MQ} & 28.53 & 46.07 & - & - \\
D&InternVideo + \textbf{EgoVideo-V} & 31.30 & 50.21 & 31.52 & 49.22  \\ 
E&InternVideo + \textbf{EgoVideo-MQ} & 31.00 & 49.28 & - & -  \\ 
F&InternVideo + \textbf{EgoVideo-V} + \textbf{EgoVideo-MQ}& \textbf{32.48} & \textbf{51.04} & \textbf{32.50} & \textbf{50.07}  \\ 
\bottomrule
\end{tabular}
\caption{The Moment Queries performance.}
\label{table:mq_result}
\end{table*}

\noindent\textbf{Approach}
We adopt ASL~\cite{asl} as our task-specific solution. 
ASL divides the task into two subtasks: classification and localization. It incorporates an action sensitivity evaluator module to assess the significance of each frame relative to the action, guiding the learning process for each subtask. The key driver for this task is Kanghua Pan.

\begin{table}[t]
\centering
\small
\setlength\tabcolsep{1.9mm}
\begin{tabular}{ccccc}
\toprule
Fast Backbone & Noun & Noun\_Verb & Noun\_TTC & Overall \\
\midrule 
X3d-M & 25.06 & 13.29 & 9.14 & 5.12 \\
\textbf{EgoVideo-V} & \textbf{31.08} & \textbf{16.18} & \textbf{12.41} & \textbf{7.21} \\
\bottomrule

\end{tabular}
\caption{Short-term object-interaction anticipation performance.}
\label{table:sta}
\end{table}

\noindent\textbf{Implementation Details.} 
\emph{1) Feature Extraction:} For further enhancing vision-only performance, we finetune the video encoder of EgoVideo-V on MQ data and the resulting model is termed as EgoVideo-MQ. Consistent with the configuration of NLQ and GoalStep, we adopt EgoVideo-V and EgoVideo-MQ to extract two types of video features.
\emph{2) Training Setup:} InternVideo, EgoVideo-V, and EgoVideo-MQ features are all projected to 512 dimensions, and other hyperparameters remain consistent with ASL. 

\noindent\textbf{Results.}
Table~\ref{table:mq_result} displays the results for MQ. Comparing \#A and \#C, our single model's features outperform the ensemble of EgoVLP and InternVideo, demonstrating the superior performance of EgoVideo. \#D combines InternVideo with EgoVideo-V features. 
Specifically, we project each feature and concatenate them. \#E incorporates InternVideo with EgoVideo-MQ features. 
\#F combines predictions from \#D and \#E by averaging the output logits for classification and localization from each model. 
Compared with ASL, our solution leverages multiple complementary features and achieves better results.

\begin{table}[t]
\centering
\small
\setlength\tabcolsep{2.8mm}
\begin{tabular}{cccc}
\toprule
Model & Noun Top1 & Verb Top1 & Action Top1 \\
\midrule 
EgoVLP & 45.53 & 40.32 & 20.63 \\
\textbf{EgoVideo-V} & \textbf{52.21} & \textbf{43.65} & \textbf{27.64}  \\ 
\bottomrule

\end{tabular}
\caption{Action recognition performance on the validation set.}
\label{table:action-val-acc}
\end{table}

\subsection{Task 4: Short-term Object-interaction Anticipation @ Ego4D}

\noindent\textbf{Task Definition}
Short-term object interaction anticipation task aims to predict the next human-object interaction happening after a given timestamp \cite{ego4d}. Given an input video, the model is required to anticipate at what time and in what location, what kind of object interaction will happen.

\noindent\textbf{Approach}
We choose to use Stillfast \cite{ragusa2023stillfast} as our downstream solution. 
This approach separately extracts high-resolution, low-frame-rate image information and low-resolution, high-frame-rate video information, and then fuses them to obtain multi-modal spatio-temporal features.
Stillfast \cite{ragusa2023stillfast} uses X3D-M \cite{feichtenhofer2020x3d} as the backbone for video feature extraction. 
We replace the X3D-M with our stronger VideoEgo-V. Differing from the original Stillfast framework which fuses multiple multi-scale intermediate layers of X3D-M (fast) and ResNet (still), we interpolate the last layer feature map of VideoEgo-V into different sizes and fuse them into the multi-scale still features generated by ResNet. The key driver for this task is Guo Chen.

\noindent\textbf{Implementation Details.} 
We adopt the training setup consistent with Stillfast. The difference is that we set the drop path rate to 0.3, and layer-wise lr decay to 0.9. Meanwhile, we enable BF16 for stable training.

\noindent\textbf{Results.}
Table~\ref{table:sta} displays the results for Short-term object-interaction anticipation on the test set. The results indicate that our EgoVideo-V is also suitable for direct transfer to forecasting tasks. In particular, the predictions of Verb and TTC are challenging to substantiate with direct evidence and often rely on advanced cognitive reasoning abilities.

\subsection{Task 5: Long-term Action Anticipation @ Ego4D}

\noindent\textbf{Task Definition}
Long-term action anticipation is a task that aims to predict multiple future actions following a given action. Each action is composed of a verb and a noun. Given an input video up to a particular timestamp, which corresponds to the last visible action, The goal is to predict a list of the twenty subsequent actions. 

\noindent\textbf{Approach}
Recent methods \cite{zhao2023antgpt, kim2023lalm} leveraging Large Language Models (LLMs) have shown superior performance in LTA tasks by converting video actions into natural language sequences, which LLMs then use to predict future actions. 
For LLM-based methods, better classification prediction and stronger LLM intuitively bring stronger language comprehension and prediction capabilities. The key driver for this task is Yicheng Liu.

\noindent\textbf{Video Clip Classification.}
Previous methods typically used video encoders like EgoVLP\cite{zhao2023antgpt, kim2023lalm} or CLIP\cite{zhao2023antgpt} combined with a Transformer-based classification head to obtain verbs and nouns. We simply finetune EgoVideo-V on LTA data to replace the previous classification predictions with our better inference results.

\noindent\textbf{Anticipation with LLMs.}
We employed the Vicuna-7B \cite{zheng2023judging} model as the LLM. During fine-tuning, we fixed the historical action sequence length to 8 and used the subsequent 20 actions as labels. 
We used EgoVLP \cite{egovlp} to extract features and augment the training set.

\noindent\textbf{Experiments}
\textbf{Implementation Details.} Following \cite{zhao2023antgpt}, during the fine-tuning phase, we set the learning rate to 3e-4, gamma to 0.85, batch size to 32, and the number of epochs to 3 for all models. We also use LoRA \cite{hu2021lora} to improve the speed and efficiency of fine-tuning.

\noindent\textbf{Action Recognition Results.} Table \ref{table:action-val-acc} shows the accuracy of action recognition on the validation set. 
The results reveal that our EgoVideo-V can achieve better prediction for the next long-term anticipation.

\begin{table*}[h!]
\centering
\small
\setlength\tabcolsep{2.6mm}
\begin{tabular}{llcccccc}
\toprule
 & & \multicolumn{3}{c}{Validation} & \multicolumn{3}{c}{Test} \\
   \cmidrule(lr){3-5} \cmidrule(lr){6-8}
LLM & Classification Model & Noun ED$\downarrow$ & Verb ED$\downarrow$ & Action ED$\downarrow$ & Noun ED$\downarrow$ & Verb ED$\downarrow$ & Action ED$\downarrow$ \\
\midrule 
LLaMA2-7B\cite{touvron2023Llama}& CLIP & 67.55 & 67.28 & 89.31 & - & - & - \\
LLaMA2-7B\cite{touvron2023Llama}& EgoVLP & 65.97 & 67.30 & 88.83 & - & - & - \\
LLaMA2-7B\cite{touvron2023Llama}& EgoVideo-V & 65.09 & 66.78 & 87.93 & 67.04 & 65.07 & 87.39 \\
Mistral-7B\cite{jiang2023Mistral}& EgoVideo-V & 65.02 & 70.08 & 88.69 & 65.00 & 68.07 & 87.70 \\
LLaMA3-8B& EgoVideo-V & - & - & - & 64.54 & 67.77 & 87.78 \\
Vicuna-13B\cite{zheng2023judging}& EgoVideo-V & 65.01 & 69.15 & 88.52 & - & - & - \\
Vicuna-7b\cite{zheng2023judging}& EgoVideo-V & \textbf{62.64} & \textbf{65.76} & \textbf{86.19} & \textbf{63.67} & \textbf{63.54} & \textbf{85.04} \\
\bottomrule
\end{tabular}
\caption{Results on the validation and test set of Long-term action anticipation Challenge.}
\label{table:lta_result}
\end{table*}

\noindent\textbf{Action Anticipation Results.} 
Table \ref{table:lta_result} shows the LTA results on the validation and testing set. The table shows that classification results of EgoVideo-V achieved significant improvements in anticipation performance compared with EgoVLP \cite{egovlp}, when using LLaMA2-7B for anticipation.
Furthermore, we tested various LLMs, including LLaMA2-7B \cite{touvron2023Llama}, LLaMA3-8B, Vicuna-7B \cite{zheng2023judging}, Vicuna-13B\cite{zheng2023judging}, and Mistral-7B \cite{jiang2023Mistral}. The Vicuna-7B demonstrated significant performance improvements.

\subsection{Task6: Action Recognition @ EPIC}
\noindent\textbf{Task definition.} Action recognition considers a short video clip and requires the model to predict the verb/noun/action classes of
the action in this segment. The evaluation metric includes Top-1/5 Accuracy. 

\noindent\textbf{Training.} 
Following prior works~\cite{avion,lavila}, we train our model for 100 epochs on the training set with a learning rate of 1e-5 and batch size of 48. 
We conduct warm-up training for 2 epochs using the cross-entropy loss. The model is trained on 16 A100 GPUs.

\noindent\textbf{Results.} Table~\ref{tab:ar} present fine-tuned model's performance on EK100 action recognition. The results reveal significant advancements with our proposed method, surpassing state-of-the-art approaches in both Verb/Noun/Action top-1 scores. 
Our single EgoVideo-V achieves 72.9\%/68.7\%/56.2\% Verb/Noun/Action top-1 scores on the test set. This is far ahead of that last challenge champion whose ensembled Verb/Noun/Action top-1 results is 71.7\%/65.8\%/54.3\%. After ensembling three different models, our EgoVideo-V further achieves slight improvement +0.2\%/+1.1\%/+0.6\%, and the final testing results are 73.1\%/69.8\%/56.8\%.
Overall, the results underscore the effectiveness of our approach in enhancing the understanding of daily human activities captured in egocentric views, highlighting its potential for advancing research in activity recognition domains.

\begin{table}[t]
\centering
\caption{Action Recognition Top-1 performance on EK100 dataset.}
\resizebox{\columnwidth}{!}{
  \begin{tabular}[t]{lcccccc}
    \toprule
     \multirow{3}{*}{Method}& \multicolumn{3}{c}{Val}&
     \multicolumn{3}{c}{Test} \\
     \cmidrule(r){2-4}  \cmidrule(r){5-7}
     
    & Verb & Noun & Action & Verb & Noun  & Action \\
\midrule
    LaViLA~\cite{lavila} & 72.0 & 62.9 & 51.0 &- &- &  - \\
    AVION~\cite{avion} & 73.0 & 65.4 & 54.4 &  - & - & - \\
    AVION~\cite{avion} (Ensemble) & - & - & - &  71.7 & 65.8 & 54.3\\
    \midrule
    EgoVideo-V & - & - & - & 72.9 & 68.7 & 56.2 \\
    EgoVideo-V (Ensemble) & - & - & - & \textbf{73.1} & \textbf{69.8} & \textbf{56.8} \\
    \bottomrule
  \end{tabular}
}
\vspace{-0.5pt}
\label{tab:ar}
\end{table}

\begin{table}[t]
\centering
\caption{Unsupervised Domain Adaptation for Action Recognition Top-1 performance on the target domain. The model is finetuned on the source domain. }
\resizebox{\columnwidth}{!}{
  \begin{tabular}[t]{lcccccc}
    \toprule
     \multirow{3}{*}{Method}& \multicolumn{3}{c}{Val}&
     \multicolumn{3}{c}{Test} \\
     \cmidrule(r){2-4}  \cmidrule(r){5-7}
     
    & Verb & Noun & Action & Verb & Noun  & Action \\
\midrule
    Previous Top-1 & - &- & - &58.2 &40.3 &  30.1 \\
    EgoVideo-V & - & - & - & \textbf{61.3} & \textbf{56.2} & \textbf{43.2} \\
    \bottomrule
  \end{tabular}
}
\vspace{-0.5pt}
\label{tab:udaar}
\end{table}

\begin{table}[t]
\centering
\caption{Zero-shot multi-instance retrieval performance on EK100 dataset.}
\scalebox{0.85}{
\begin{tblr}{
  cells = {c},
  hline{1-2,8} = {-}{},
}
Method  & Backbone   &  Average mAP & Average nDCG  \\
EgoVLP~\cite{egovlp} & ViT-B & 16.6 & 23.1 \\
LaViLA~\cite{lavila} & ViT-B & 30.9 & 32.0 \\
AVION~\cite{avion} & ViT-B & 30.9 & 32.0 \\
LaViLA~\cite{lavila} & ViT-L & 36.1 & 34.6 \\
AVION~\cite{avion} & ViT-L & 37.6 & 35.3 \\
\hline
EgoVideo  & EgoVideo-1B & \textbf{47.6} & \textbf{39.4} \\
\hline
\end{tblr}}
\label{tab:zeroshot}
\end{table}

\begin{table}[t]
\centering
\caption{Multi-instance retrieval performance on EK100 dataset.}
\resizebox{\columnwidth}{!}{
  \begin{tabular}[t]{lcccccc}
    \toprule
     \multirow{2}{*}{Method}& \multicolumn{3}{c}{mAP}&
     \multicolumn{3}{c}{nDCG} \\
     \cmidrule(r){2-4}  \cmidrule(r){5-7}
     
    & Avg. & T2V & V2T & Avg. & T2V  & V2T \\
    \cmidrule(r){1-1}  
    \cmidrule(r){2-4}  \cmidrule(r){5-7}
    EgoVLP~\cite{egovlp} & 45.0 & 40.5 & 49.9 & 59.4 & 57.9 & 60.9 \\
    LaViLA~\cite{lavila} & 50.9 & 47.1 & 54.7 & 66.5 & 64.9 &  68.1 \\
    AVION~\cite{avion} & 54.5 & 51.1 & 57.9 & 69.0 & 67.6 & 70.4 \\
    \hline
    EgoVideo & \textbf{63.3} & \textbf{58.9} & \textbf{67.6} & \textbf{73.2} & \textbf{71.5} & \textbf{75.0} \\
    \bottomrule
  \end{tabular}
}
\label{tab:finetune}
\end{table}

\subsection{Task7: Multi-instance Retrieval @ EPIC}
\noindent\textbf{Task definition:} 
The primary objective of Epic-Kitchen Multi-Instance Retrieval task is to develop models capable of accurately retrieving relevant video segments from the Epic-Kitchen-100 dataset given a query in the form of a textual description of the action or activity.
The evaluation metric includes Mean Average Precision (mAP) and normalized Discounted Cumulative Gain (nDCG). More detailed information can be found in~\cite{ek100}. 

\noindent\textbf{Training:} 
Following prior works~\cite{avion,lavila}, we train our model for 50 epochs on the training set with a learning rate of 1e-5 and batch size of 8. We conduct warm-up training for 1 epoch using the classic video-text contrastive loss. The model is trained on 8 A100 GPUs for 12 hours. 

\noindent\textbf{Results.} Tables~\ref{tab:zeroshot} and ~\ref{tab:finetune} present zero-shot and fine-tuned model's performance on EK100 multi-instance retrieval. 
Comparative analysis revealed significant advancements with our proposed method, surpassing state-of-the-art approaches in both mAP and nDCG scores. 
As shown in Table~\ref{tab:zeroshot}, the zero-shot performance of our stage 2 model (after post-training) reveals strong retrieval performance, compared with EgoVLP and LaViLA, indicating the strong performance of our backbone model and the effectiveness of the multi-stage training strategy.
Through task-specific training, our model achieves 63.3\% and 73.2\% average mAP and nDCG, respectively, exhibiting substantial improvements in both text-to-video and video-to-text retrieval tasks. This indicates superior performance in capturing fine-grained action semantics within the kitchen domain. 
Overall, the results underscore the effectiveness of our approach in enhancing the understanding of daily human activities captured in egocentric views, highlighting its potential for advancing research in activity recognition and video retrieval domains.

\subsection{Task8: Domain Adaptation for Action Recognition @ EPIC}
\noindent\textbf{Task definition.} Domain Adaptation is defined by 
utilizing a labelled source domain to train an action recognition model that is capable of adapting to
an unlabelled target domain. According to the data source~\cite{ek100}, this task poses additional challenges due to the discrepancy in location, hardware, and long-term temporal
offsets. The evaluation metric includes Top-1/5 Accuracy. 

\noindent\textbf{Training.} 
Similar to the training setting of action recognition, our approach differs in that we only train the model on the source domain.

\noindent\textbf{Results.} Tables~\ref{tab:udaar} present model's performance on EK100 domain adaptation action recognition. Notably, our model is only finetuned on the source domain, achieving 61.3\%/56.2\%/43.2\% Verb/Noun/Action top-1 performance that is much higher than the previous leading results 58.2\%/40.3\%/30.1\%. This highlights superior performance improvement brought by well-pretrained models. 

\section{Limitation and Conclusion}
Although our solution achieved good results in the competition, there are still some limitations worth noting. Firstly, we use a large video-language model and A100 as the computing GPU during the training process, which requires expensive computing resources and results in higher carbon emissions. Secondly, we employ feature-based approaches to solve the temporal localization problem, which often fails to obtain the optimal solution. Finally, we find that in Long-Term Action Anticipation (LTA) tasks, training and prediction based on LLMs have high uncertainty, and the final prediction performance may not be proportional to the capability of the LLM itself.

In conclusion, we have presented our solutions to 8 tracks in the
EgoVis CVPR2024 Challenge. We find a larger video-language model can still give an advantage to egocentric task performance. This reveals that there is still ample room for exploration in egocentric video understanding.

{\small
\bibliographystyle{ieee_fullname}
\bibliography{egbib}
}

\end{document}